%% file: main.tex
\documentclass[final,12pt]{elsarticle}

\usepackage{hyperref}
\hypersetup{
     colorlinks = true,
     linkcolor = blue,
     anchorcolor = blue,
     citecolor = blue,
     filecolor = blue,
     urlcolor = blue
     }

\usepackage{graphicx}
\usepackage{bm}
\usepackage{amsfonts} 
\usepackage{lscape}

\usepackage[utf8]{inputenc}
\usepackage{pgfplots}

\DeclareUnicodeCharacter{2212}{−}
\usepgfplotslibrary{groupplots,dateplot}
\usetikzlibrary{patterns,shapes.arrows}
\pgfplotsset{compat=newest}

\pgfplotsset{every axis/.append style={
        scaled y ticks = false, 
        scaled x ticks = false, 
        y tick label style={/pgf/number format/.cd, fixed, fixed zerofill,
                            int detect,1000 sep={\;},precision=3},
        x tick label style={/pgf/number format/.cd, fixed, fixed zerofill,
                            int detect, 1000 sep={},precision=3}
    }
}

\usepackage{xspace}
\newcommand{\system}{\textsc{ModularHI}\xspace}  
\newcommand{\supervisor}{aggregator\xspace}  
\newcommand{\Supervisor}{Aggregator\xspace}  
\newcommand{\niceHI}[1]{\ensuremath{\textrm{HI}^{#1}}\xspace}   
\newcommand{\HIcombo}{\niceHI{j}} 
\newcommand{\HImodel}{\niceHI{m}}

\usepackage{caption}
\usepackage{subcaption}

\usepackage{url}

\usepackage[ruled,vlined]{algorithm2e}

\journal{arXiv}

\begin{document}

\begin{frontmatter}
\title{A data-driven modular architecture with denoising autoencoders for health indicator construction in a manufacturing process}







\author[aauaddr,ucnaddr]{Emil Blixt Hansen\corref{mycorrespondingauthor}}
\cortext[mycorrespondingauthor]{Corresponding author, Fibigerstræde 16, 9220 Aalborg East, Denmark}
\ead{ebh@mp.aau.dk}

\author[ntnuaddr]{Helge Langseth}

\author[ucnaddr]{Nadeem Iftikhar}

\author[aauaddr]{Simon Bøgh}

\address[aauaddr]{Department of Materials and Production, Aalborg University, Denmark}
\address[ucnaddr]{Industrial Digital Transformation, University College of Northern Denmark, Aalborg, Denmark}
\address[ntnuaddr]{Department of Computer Science, Norwegian University of Science and Technology, Trondheim, Norway}

\begin{abstract}
Within the field of prognostics and health management (PHM), health indicators (HI) can be used to aid the production and, e.g. schedule maintenance and avoid failures. However, HI is often engineered to a specific process and typically requires large amounts of historical data for set-up. This is especially a challenge for SMEs, which often lack sufficient resources and knowledge to benefit from PHM. In this paper, we propose \system, a modular approach in the construction of HI for a system without historical data. With \system, the operator chooses which sensor inputs are available, and then \system will compute a baseline model based on data collected during a burn-in state. This baseline model will then be used to detect if the system starts to degrade over time. We test the \system on two open datasets, CMAPSS and N-CMAPSS. Results from the former dataset showcase our system's ability to detect degradation, while results from the latter point to directions for further research within the area. The results shows that our novel approach is able to detect system degradation without historical data.\end{abstract}


\end{frontmatter}

\section{Introduction}
The manufacturing industry has seen an increased interest in modern IT technologies to enhance their production. 
These technologies are commonly described as Industry 4.0. Nonetheless, many of the processes use old equipment, which often lacks sensors and communication protocols to be used with technologies of Industry 4.0, such as big data and analytics \cite{Gurkov2007}. A study also found that the use of old equipment negatively impacts the competitiveness and innovation of a small and medium-sized enterprises (SME) \cite{Abrham2015}. Prognostics and health management (PHM) concerns the topic of monitoring and health indication of machinery. The health indicator (HI) part is concerned with estimating the health of machinery. The HI can be constructed differently depending on the equipment and setting, and an important design choice is whether the HI is to be model-based, data-driven, or hybrid \cite{oatao19521}. Model-based models takes the system's underlying information into accounts using mathematical models or other descriptions of the system behaviour. The data-driven approach is where the HI score is not built on top of system information, but rather learned from data. Lastly, the hybrid approach is a combination of the other two. In this work we focus on the \textit{data-driven} approach. Within this category, there are different ways the HI can be constructed, and examples include  signal processing and statistical techniques, including time series domains using time-domain or frequency-domain feature \cite{Atamuradov2020,zhu2014survey}. The continued development in deep learning has resulted in several approaches to construct HI through deep learning. A variant of autoencoders (AE) is for instance commonly used \cite{8294247,ZHAO2019213}. Experiments from Zhao et al. \cite{ZHAO2019213} showed that denoising autoencoders (DAE) outperformed other methods for multivariate time series reconstruction problems, whereas \cite{GUO2018142} constructed a convolutional neural network (CNN) to construct the HI directly.

Within the PHM domain, estimating the remaining useful life (RUL) of a given machinery can be beneficial. RUL estimation is often based on the HI score, but the RUL estimates will then typically rely on historical data of failures to estimate the relationship between HI and RUL. Ensemble learning with time dependant degradation weights has been shown to generate accurate RUL predictions \cite{LI2019110}, while \cite{BERGHOUT2020103936,Saxena2008} used an adaptive denoising algorithm to make feature extraction and compute the RUL of aircraft engines. Moreover, it has been shown that utilising semi-supervised learning can improve the performance of RUL prediction \cite{LISTOUELLEFSEN2019240}. A combination of ARIMA and LSTM models has been proposed to increase performance in anomaly detection and forecasting \cite{Hollingsworth2018}. 

The use of HI for PHM is often engineered to the specific machine/task, and furthermore requires that sufficient data of the machine is available to construct the HI. However, SMEs does not in general have this available to them; when it comes to Industry 4.0, they are often behind larger enterprises and do not have the knowledge to use technologies such as artificial intelligence (AI) \cite{HANSEN2021362}. Previous work has focused on how to make these technologies more available for SMEs, focusing on ease of use. For instance, Hansen et al.\ \cite{HANSEN20201146} described a framework and conducted tests within image classification problems. 
 
To make HI scoring easy-to-use for SMEs, it is not feasible to assume that the HI scoring system is particularly engineered to each specific machine used by the SME. On the contrary, it should be possible for an operator to set up the framework without understanding the underlying algorithms. Furthermore, since a production process typically consists of multiple different machinery and equipment, it could be beneficial for the SME not to be limited to a specific engineering application. Therefore, \system\ is not limited to specific sensor input types. Instead it uses a modular divide-and-conquer strategy, which is in-line with current thinking in AI. For instance, DeepMind \cite{fernando2017pathnet} presented PathNet, a scalable modular network suited for transfer learning between different tasks in 2017. Here small neural network models together form a larger network. The work was later extended by Stepwise PathNet \cite{Imai2020}. These works demonstrate that modularity play a central role within deep learning to enhance the overall performance.

In this paper we present \system, a novel approach to construct an HI score for an arbitrary machinery based on the available sensor inputs. \system is aimed at, but not limited to, use cases within the production sector. Our working hypothesis is  that while it is infeasible to engineer health monitoring solutions for each machine, it is still beneficial to monitor a machine's health even if the employed system is less accurate than a tailor-made solution would have been. The goal is thereby to provide a low-cost and generally applicable HI monitoring system for SMEs, that can enable them to start benefitting from Industry 4.0 technologies. The rest of the paper is structured as follows. In \autoref{sec:system architecure} the algorithm and components of \system is explained. In \autoref{sec:experiments} two different tests scenarios are described and in \autoref{sec:results} the results are shown. Furthermore, in \autoref{sec:discussion} and \autoref{sec:conclusion} the discussion and conclusion is presented, respectively. 

\section{System architecture} \label{sec:system architecure}
\begin{figure*}[ht]
    \centering
    \includegraphics[width=\textwidth]{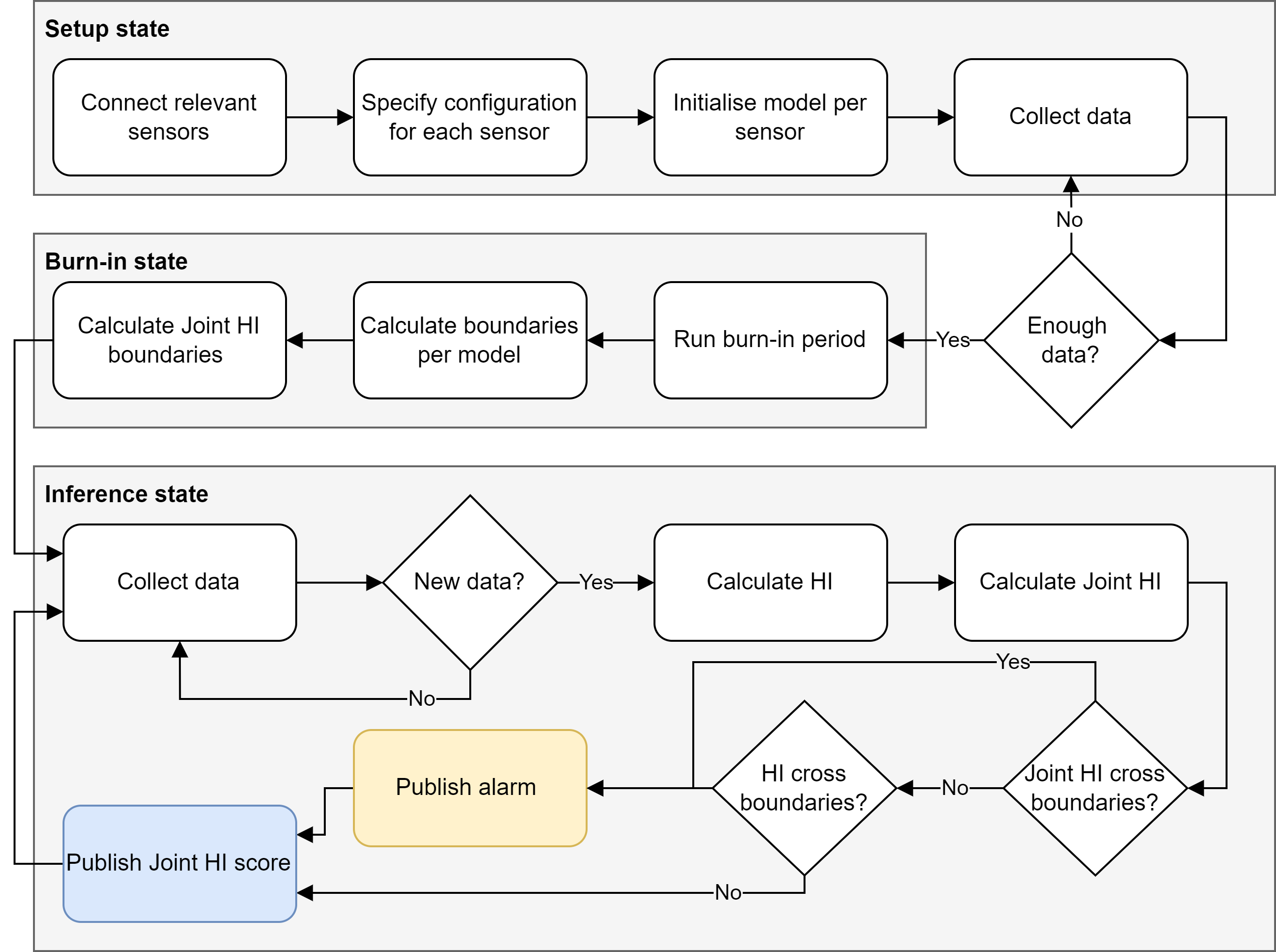}
    \caption{The execution flow of the three states: setup, burn-in, and inference.}
    \label{fig:execution flow}
\end{figure*}

\system aims to support a broad set of use cases, and is not engineered for specific tasks. We have therefore split the execution into three states:  \textit{setup}, \textit{burn-in}, and \textit{inference}. The general execution flow of \system's three states is shown in \autoref{fig:execution flow}. Besides the execution flow, the underlying system comprises two main parts: the \textit{component models} and the \textit{\supervisor}. The execution flow and the two main parts are discussed next. 

\subsection{Execution flow} \label{subsec:execution flow}
In the first state, \textit{setup state}, the different models and their specific input are configured. The models are called \textit{component models}, and are further described in \autoref{subsec:component models}. Furthermore, the initial data collection also takes place in this state. When enough data is collected, the execution goes to the burn-in state. 

The \textit{burn-in state} is where the component models are fitted to the collected datasets. We will assume that the data describes the monitored system in its optimal performance state, preferably directly after machine maintenance. When each component model has been sufficiently trained on their individual datasets, the components model can calculate a real-time health indicator (more on this in \autoref{subsec:component models}). In the inference state the component model will monitor its own status, and posts an alarm if the HI value is deemed to be outside its acceptable region. First, however, the acceptable region must be established. While several methods can be used to determine this acceptable region, we use a simple strategy in this paper. We monitor the HI during the burn-in state, calculate the standard deviation of the calculated HI during this period, and define the acceptable region for the HI to between zero and nine times the calculated standard deviation. Finally we define the HI at the system level. This is done by the \supervisor model, which collects and combines the HI values from the component models. The combination is named the \textit{joint HI} (\HIcombo), represents the overall system's HI score, and has its own acceptable region. Again, the region is defined by nine standard deviations calculated during the burn-in stage. The combinations of the different component models is further described in \autoref{subsec:supervisor}. Next, the execution enters the third state, the inference state.

The \textit{inference state} is where the \system is  executed on new data. Here a continuous collection of data occurs, and every time a new data point is collected, it is executed through the models. This includes calculating the current HI score comparing the new datapoint to data collected during the burn-in state. After that, $\HIcombo$ is calculated, and all of the boundaries are checked. If at least one boundary is crossed, the system is assumed to either be in a faulty state or sliding towards it, indicating that maintenance or inspection is recommended. This therefore results in an alarm state being published.

\subsection{Component models} \label{subsec:component models}
\system has been designed with simplicity of use in mind. To handle different sensor categories, with different data types and data dimensions, we therefore designed specific models per sensor type. Hence, each sensor has its own model, specifically designed for that measurement type. It follows that component models are specifically trained for a certain type of measurement (e.g., a specific model for temperature data). 

As mentioned above, there are different ways of constructing a HI score. One of the requirements for \system is that it should be modular and handle a different mixture of sensor inputs. Therefore, it is infeasible to train a separate model to generate the system's HI score for each possible scenario. Instead, \system consists of individual component models, each outputting a HI score based on a single stream of data. To enable these HI scores to be comparable to each other throughout the different sensor inputs, we have chosen the DAE \cite{ZHAO2019213} as our go-to model type. Since the component models are designed to only handle univariate time series, the model size is limited. In the experiments reported in this paper, we have chosen to use a DAE with LSTM layers to capture the time perspective in the data. The DAE architecture is the same across all sensors and all experiments, and can be seen in \autoref{tab:base dae}. We note that the same core modelling approach could be used to analyse other streaming data like sound or video simply by adapting the DAE to handle that data-type. 

\begin{table}[htb]
\centering
\caption{The LSTM autoencoder.}
\begin{tabular}{l|l}
\textbf{Layer type} & \textbf{Specification} \\
\hline
Input               &   Shape (batch size, window size, 1) \\
LSTM                &   8 units, returned sequence  \\
Dropout             &   Probability 0.5 \\
LSTM                &   4 units \\
Repeat vector       &   8 times \\
LSTM                &   4 units, returned sequence  \\
Dropout             &   Probability 0.5 \\
LSTM                &   8 units, returned sequence  \\
Output              &   Time distributed of window size \\
\end{tabular}
\label{tab:base dae}
\end{table}

Currently, this DAE architecture is used for all time series sensor input, such as temperature, vibration, etc. Even though all models have the same architecture, they are independently pre-trained on a relevant dataset. This application of transfer-learning reduces the data-requirements during system setup, reduces the compute required for training the model during burn-in, and improves the model's performance \cite{9134370}. 

The HI score is based on the mean absolute error (MAE) during reconstruction of the input-signal:
\begin{equation}\label{eq:MAE}
\HImodel_k=\frac{1}{n} \sum_{i=1}^{n}\left|x_{i}-\tilde{x}_{i}\right|.
\end{equation}
Here $\HImodel_k$ refers to the HI score of the $k$'th component model $m$ (out of the set of $M$ component models for a specific setup case), $\bm{x}$ is a vector of measurements of window-size $n$ and $\bm{\tilde{x}}$ is the corresponding reconstructed measurements also of size $n$. As stated earlier, each model has its own upper-bound calculated by taking the mean of the HI of burn-in and finding standard deviation $\sigma$ from the sample mean. Specifically, the upper boundary is specified by $9\,\sigma$ and the lower is set to 0 (since MAE is a non-negative real number).

\subsection{\Supervisor} \label{subsec:supervisor} 
The objective of the \textit{\supervisor} is to combine the information from the component models to give a system-wide HI, which we call the \textit{joint HI} ($\HIcombo$). We note that \HIcombo can be calculated in many different ways, ranging from simple aggregations like maximum or sum to increasingly complex combinations, e.g., represented by a deep neural network. As \system is designed for ease-of-use within a wide range of application areas,  to be data-driven, and to not require engineering input to define the aggregation function, we choose to let each sensor contribute equally to \HIcombo  by default. The joint health indicator is now simply the average of the $N$ components' HI scores:

\begin{equation}\label{eq:default joint HI}
\HIcombo=\frac{1}{N}\sum_{i=1}^{N} \HImodel_i
\end{equation}

Nevertheless, if an operator knows that some sensor is more critical to the stability of a system, they can specify weights $\bm{w}$ for each sensor input. An example of this could be that the operator knows that the system starts to be unstable when the system temperature rises. The operator will then give the temperature sensor input a higher weight the the others. Thus the $\HIcombo$ will be more acceptable to changes from the temperature sensor. The $\HIcombo$ is then calculated by a weighted average as seen in \autoref{eq:joint HI}. By default all the weights are specified as 0.5, hence they are all equal and the \HIcombo will be calculated as \autoref{eq:default joint HI}.

\begin{equation}\label{eq:joint HI}
\HIcombo=\frac{\sum_{i=1}^{N} w_{i} \cdot \HImodel_i}{\sum_{i=1}^{N}{w_{i}}}
\end{equation}

\system calculates the upper-bound for \HIcombo at $9\,\sigma$  during the burn-in state. This is done when all component models have finished their burn-in step. Then \autoref{eq:joint HI} is calculated to find all of the \HIcombo from the burn-in period to calculate the sample mean and thus the corresponding $\sigma$. 

\begin{figure*}[ht]
    \centering
    \includegraphics[width=0.7\textwidth]{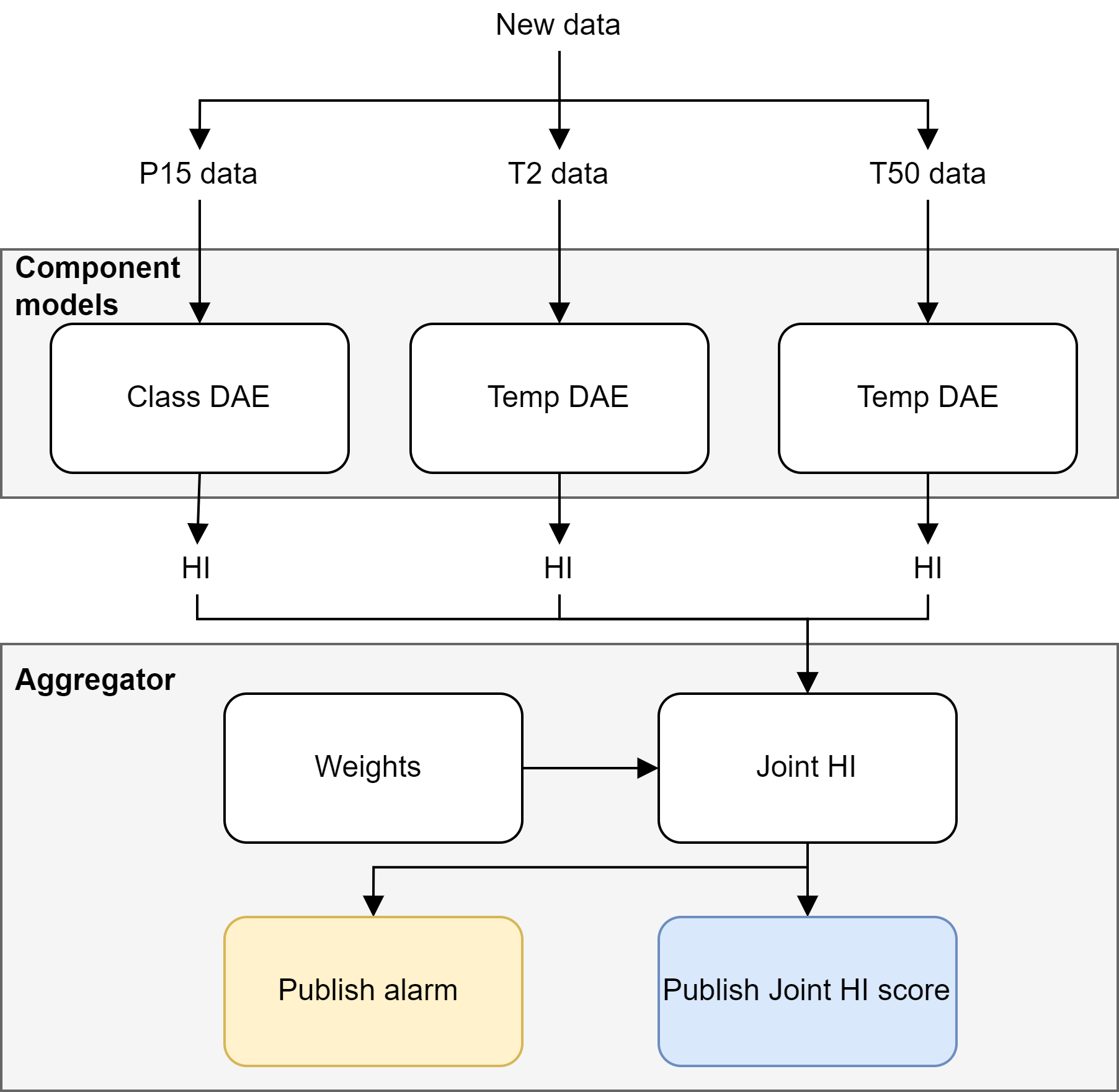}
    \caption{The combination of the different component models and the \supervisor from CMAPSS T7.}
    \label{fig:modularhi c t7}
\end{figure*}

\begin{sloppypar}
When the burn-in state is completed, the cycle of the inference state is executed. For a new data point $x_{t}$ all component models will run a window of the last $n$ samples including the new data point, such that ${\bm{x} = \{x_{t - n + 1}, \ldots, x_{t}\}}$. After that, all calculated $\HImodel$ values will be used to calculate the current $\HIcombo$. With the new $\HIcombo$ all boundaries are checked such that if any boundary is crossed in their respected HI, an alarm is published. An illustration of the inference state of \system can be seen in \autoref{fig:modularhi c t7} where test 7 from \autoref{sec:cmapss dataset} is shown. 
\end{sloppypar}

\section{Experiments} \label{sec:experiments}
To validate the use of \system, we needed to test it on a relevant dataset. This dataset should preferably contain multiple different sensors (to validate the modularity), include recordings from a steady-state (to enable the initial data collection), and have a continuous data steam eventually leading to a faulty state. Finally, it would be beneficial if the dataset was relevant to production applications, either being extracted from one or similar to a process that could exist in one. The MIMII dataset, which contains various recordings of failure of processes \cite{Harsh2019MIMII} is one potential candidate. However, it does not contain continuous recording from steady-state to failure, and does not include information such as vibration. A dataset collected from a real production line containing various different sensors and faulty states was presented in \cite{Hansen2022AICD}. Unfortunately, this dataset consists mainly of operations failures and not process failures. NASA published the jet engine degradation dataset CMAPSS in 2008 \cite{saxena2008turbofan}. This dataset contains simulated jet engines monitored until failure, where each data point is an aggregated value from one flight. More recently, they released an updated version, named N-CMAPSS \cite{Chao2021Aircraft}. This dataset contains various sensor values from the same setting. The main difference is that data in N-CMAPSS is collected throughout the flights and includes both ascent and descent. Moreover, the initial states of the flights are collected at actual flights, and only the run to failure is simulated. Hence, N-CMAPSS is a larger and more complex dataset compared to the original CMAPSS. We have chosen to test \system against both CMAPSS and N-CMAPSS. We chose both datasets because both of them have a continuous data stream from steady-state until failure. Moreover, the two datasets symbolise two different levels of complexity. The available datasets also show the lack of available authentic production datasets which can be used in our case. Thus the two CMAPSS datasets are the closest to an actual production dataset with their continuous data stream of various sensors. 

\subsection{CMAPSS dataset test} \label{sec:cmapss dataset}
The original CMAPSS dataset consists of multiple flights, which are commonly used for RUL predictions \cite{BERGHOUT2020103936,Saxena2008}.  Here, however, we use  \system to generate a HI and determine if the equipment/process is in steady operation. Therefore, our test will only consist of a single jet engine (engine number 1 from the dataset FD001). Besides which sensors to use, we also need to decide on two hyperparameters: the duration of the burn-in period and the window size used by the autoencoder. Since engine number 1 only has 220 recordings, a burn-in period of 78 is chosen as it includes a stable amount of data. A window size of 8 was chosen as it should capture enough of the temporal data. We remark that the relatively small dataset used during burn-in is sufficient due to the auto-encoder being pre-trained on a separate dataset. The component model concerning temperature is pre-trained on historical weather data\footnote{\url{https://www.kaggle.com/budincsevity/szeged-weather}} and the component models concerning generic time series data is pre-trained on accelerometer data collected at Aalborg University. The general purpose of pre-training the networks is to train them to reproduce the input; the underlying characteristics of the individual sensor data will then be learned during the burn-in state. 

\input{tabs/cmapss-test-table}

\begin{figure*}[h]
    \centering
    \includegraphics[width=\textwidth]{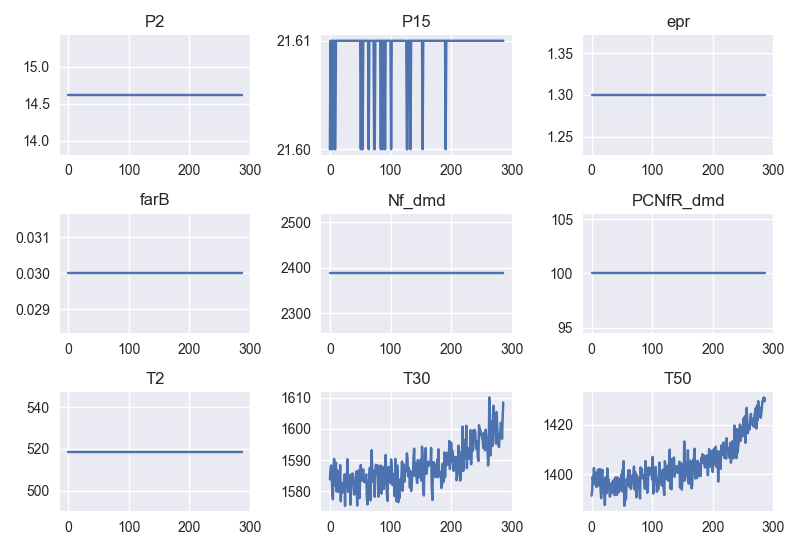}
    \caption{The nine used sensor from CMAPSS engine number 1 used, only in sensor T30 and T50 is the degradation visible with an upward trend.}
    \label{fig:cmapss-used-values}
\end{figure*}

We conducted eight tests on the data from Engine number 1, each with a different combination of sensors and weights (refer to \autoref{eq:joint HI}). The test-cases are described in \autoref{tab:cmaps test setup}. Moreover, the values for the sensors are plotted in \autoref{fig:cmapss-used-values}. The eight tests are conducted to validate various aspects of \system. Tests 1 and 2 examine the usability of \system with inputs where degradation is present. Test 5 considers what happens when degradation is not clearly present. Tests 3, 6 and 8 look at a combination of sensor with ``informative'' and ``uninformative'' sensor values with equal weights. Finally, Tests 4 and 7 (inference state visualised in \autoref{fig:modularhi c t7}) highlight the same when a larger weight has been applied to the ``informative'' measurements.   

\subsection{N-CMAPSS dataset test} \label{sec:ncmapss dataset}
Besides the original CMAPSS data, we also tested \system on the more complex N-CMAPSS dataset. Since N-CMAPSS consists of multiple flights of different lengths, we chose flight number 2 from dataset DS01. These flights comprised of data from flights over 3000 feet. All the flights include ascend from 3000 feet, cruise altitude and descending to 3000 feet. Flight number 2 describe class three flights, meaning all flights is over 5 hours long \cite{Chao2021Aircraft}. The sensor readings and flight settings are subject to high variance during a flight's ascend and descend. Since the focus of \system is to be able to detect deviations from stable operation, we have filtered out the ascending and descending parts of the flights. Moreover, only cruise parts between 25,000 and 30,000 feet were included during the burn-in state. Furthermore, we only included flights where the cruise part had a minimum of $1024$ observations (as we used window-size $n=1024$ for the autoencoders). The cleaned file consists of 4 sensors, and the first 105,876 readings are used for the burn-in state. 

\input{tabs/ncmapss-test-table}

\begin{figure*}[h]
    \centering
    \includegraphics[width=\textwidth]{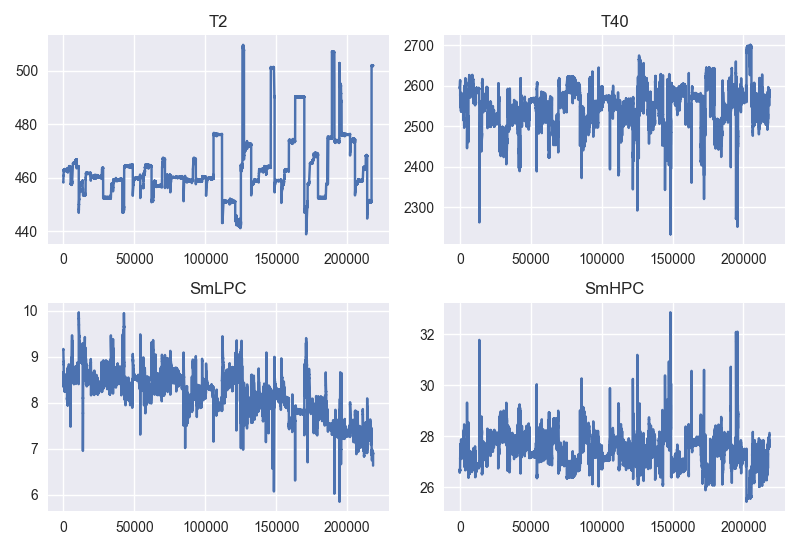}
    \caption{The 4 used sensor from N-CMAPSS engine number 2. Only in sensor SmLPC is the degradation clearly visible with an downwards trend.
    }
    \label{fig:ncmapss-used-values}
\end{figure*}

We conducted five tests of the cleaned DS01 engine 2 data. All of them had a sliding window size of 1024 and was tested in batches of 256. To combat the abrupt changes in the values when going from a single cruise to another, we only trained with mini-batches where one cruise was involved. The different tests, along with their models and weights are reported in \autoref{tab:ncmaps test setup}. A plot of the used sensors after data cleaning can be seen in \autoref{fig:ncmapss-used-values}.  While some of the sensors in  \autoref{fig:ncmapss-used-values} show signs of degradation (see, in particular, SmLPC), the apparently noisy variations in the readings dominate the signals. The dataset is therefore particularly challenging for  \system, where we for simplicity assume that all data during the burn-in period is from stable operation of the equipment/process. From just plotting the dataset, one should therefore expect that the monitoring system will be more challenged by this dataset than from the previous one. 

As with the CMAPSS dataset, the five tests were conducted to test different aspects of \system. Tests 1 and 2 was to see the effect of using only a single sensor, where one of them (Test 2) had a clear degradation. Tests 3 and 4 use multiple sensor values, and Test 5 employs sensors with different weights. 

\section{Results} \label{sec:results}
%

The component models were learning until there was no improvement in the validation loss for 25 consecutive epochs, and at that time the weights from the epoch with the lowest validation loss was used.

\subsection{CMAPSS dataset result} \label{sec:cmapss dataset result}
\begin{figure*}
     \centering
     \begin{subfigure}[b]{0.45\textwidth}
         \centering
         \includegraphics[width=\textwidth]{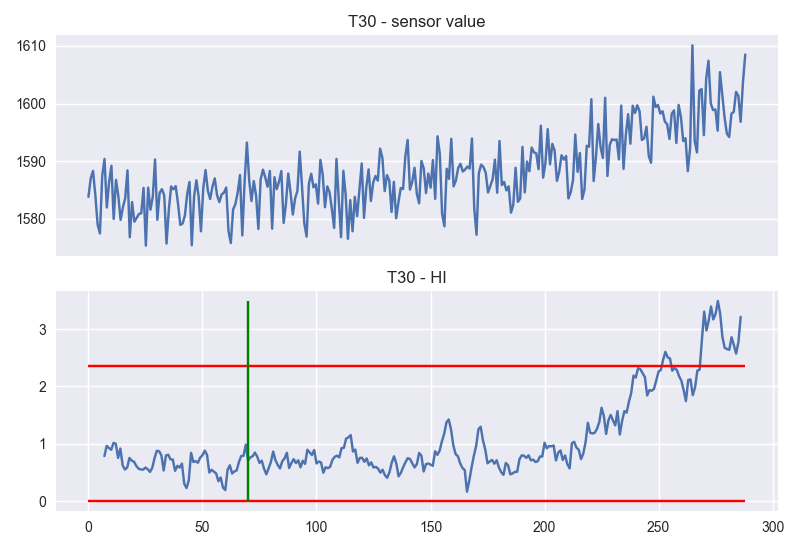}
         \caption{T30 sensor value and HI.}
         \label{fig:r-cmapss-test2-t30}
     \end{subfigure}
     \hfill
     \begin{subfigure}[b]{0.45\textwidth}
         \centering
         \includegraphics[width=\textwidth]{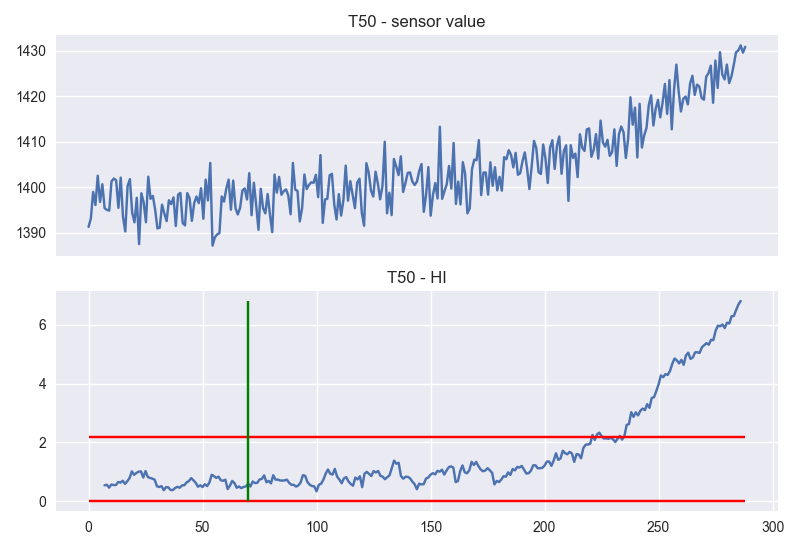}
         \caption{T50 sensor value and HI.}
         \label{fig:r-cmapss-test2-t50}
     \end{subfigure}
     \hfill \\
     \begin{subfigure}[b]{0.90\textwidth}
         \centering
         \includegraphics[width=\textwidth]{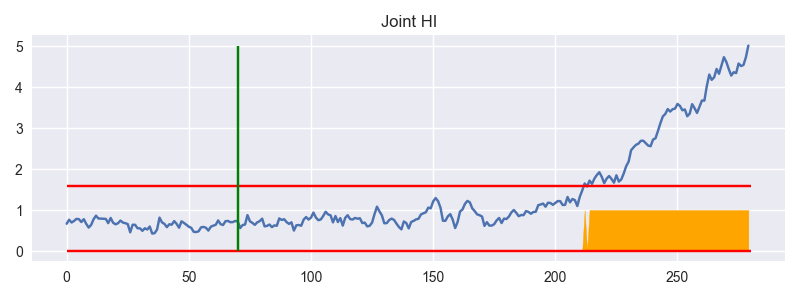}
         \caption{$\HIcombo$ for T30 and T50.}
         \label{fig:r-cmapss-test2-joint}
     \end{subfigure}
        \caption{The result from Test 2. The green vertical line indicates the end of the burn-in state. The horizontal red lines indicate the calculated boundaries. The orange area in the $\HIcombo$ is when an alarm is published.}
        \label{fig:r-cmapss-test2}
\end{figure*}

The first two tests, Tests 1 and 2, examined the system when degradation is visually present in the sensor readings. Since \textit{T50} is included in both tests, the results are almost identically. The result from Test 2 reported in \autoref{fig:r-cmapss-test2} show that $\HIcombo$ reacts at time-step 212. The two component models first publish alarms at step 221 and 252 for sensor T50 and T30, respectively. 

\begin{figure*}
     \centering
     \begin{subfigure}[b]{0.45\textwidth}
         \centering
         \includegraphics[width=\textwidth]{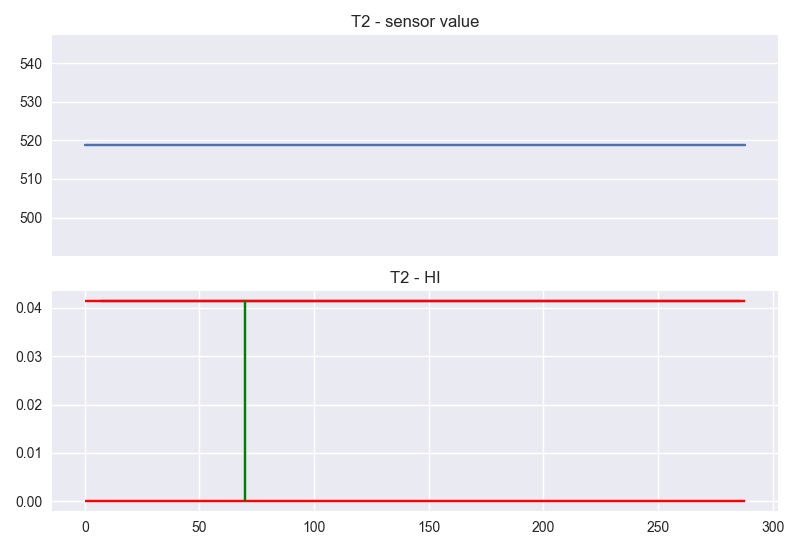}
         \caption{T2 sensor value and HI.}
         \label{fig:r-cmapss-test7-t2}
     \end{subfigure}
     \hfill
     \begin{subfigure}[b]{0.45\textwidth}
         \centering
         \includegraphics[width=\textwidth]{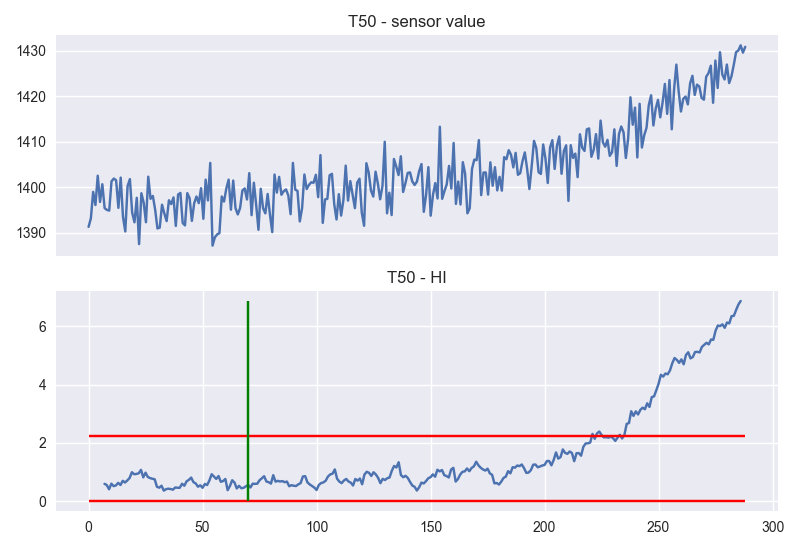}
         \caption{T50 sensor value and HI.}
         \label{fig:r-cmapss-test7-t50}
     \end{subfigure}
     \hfill \\
     \begin{subfigure}[b]{0.45\textwidth}
         \centering
         \includegraphics[width=\textwidth]{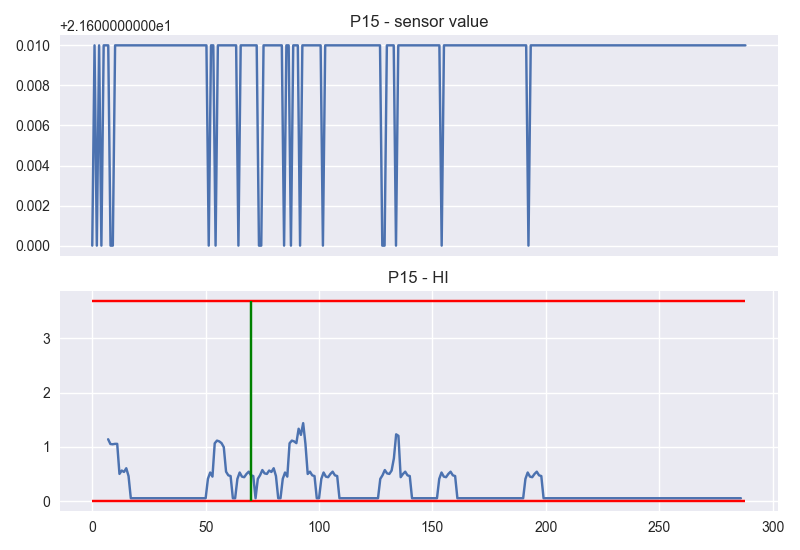}
         \caption{P15 sensor value and HI.}
         \label{fig:r-cmapss-test7-p15}
     \end{subfigure}
     \hfill
      \begin{subfigure}[b]{0.45\textwidth}
         \centering
         \includegraphics[width=\textwidth]{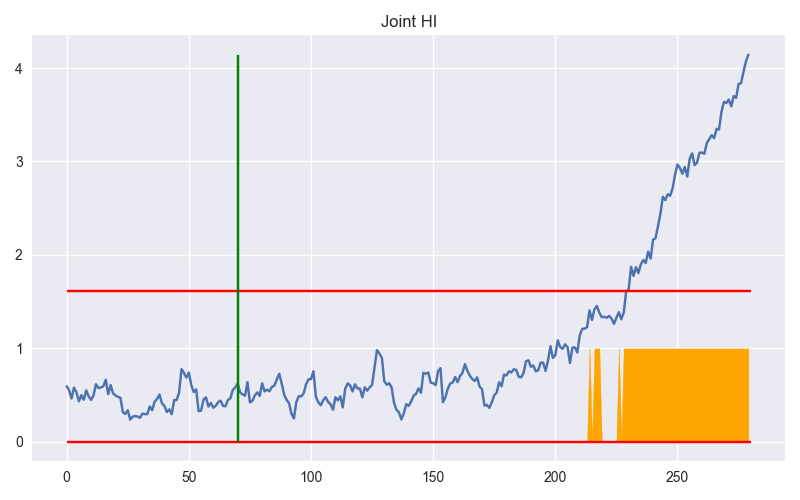}
         \caption{$\HIcombo$ for T2, T50, and P15.}
         \label{fig:r-cmapss-test7-joint}
     \end{subfigure}
        \caption{The result from Test 7. The green vertical line indicates the end of the burn-in state. The horizontal red lines indicates the calculated boundaries. The orange area in the $\HIcombo$ is when an alarm is published. The HI score from \autoref{fig:r-cmapss-test7-t2} is below the upper boundary line.}
        \label{fig:r-cmapss-test7}
\end{figure*}

In Test 7, we used sensors T50, T2, and P15. Sensor T50 displays a clear degradation in the sensor values, while sensor P15 only has a small alteration in the data, and T2 is static. The results can be seen in \autoref{fig:r-cmapss-test7}, and show that sensor T2 and P15 do not indicate any depredations, neither does their HI score. However, since sensor T50 crosses its upper-bound an alarm is published. We also note that $\HIcombo$ exceeds its upper bound even though two sensors do not indicate a degradation.

\begin{figure*}
    \centering
    \includegraphics[width=\textwidth]{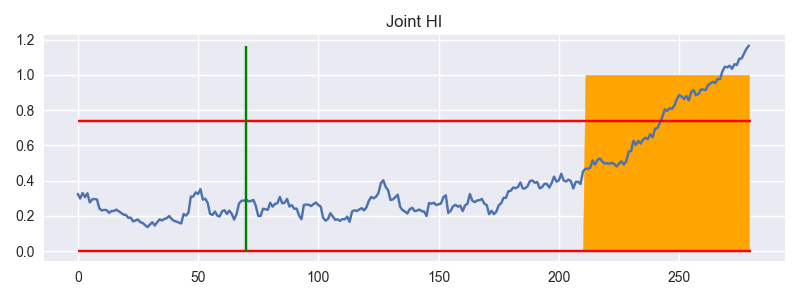}
    \caption{The result from Test 8. The green vertical line indicates the end of the burn-in state. The horizontal red lines indicates the calculated boundaries. The orange area in the $\HIcombo$ is when an alarm is published.}
    \label{fig:r-cmapss-test8}
\end{figure*}

Test 8 was conducted to see if a majority of non-contributing sensor values will prevent $\HIcombo$ to detect that one sensor measures a degradation. Therefore, this test includes one sensor with the degradation evident (T50) and six  without (P2, P15, epr, farB, Nfdmd, and PCNfRdmd). The $\HIcombo$ results in \autoref{fig:r-cmapss-test8} show that an alarm is published before $\HIcombo$ crosses its boundary. This is initiated by the sensor T50 crosses its upper-bound. Moreover, it can also be seen that $\HIcombo$ itself crossed its boundary at time 243. 

\subsection{N-CMAPSS dataset result} \label{sec:ncmapss dataset result}
\begin{figure*}
     \centering
     \begin{subfigure}[b]{0.45\textwidth}
         \centering
         \includegraphics[width=\textwidth]{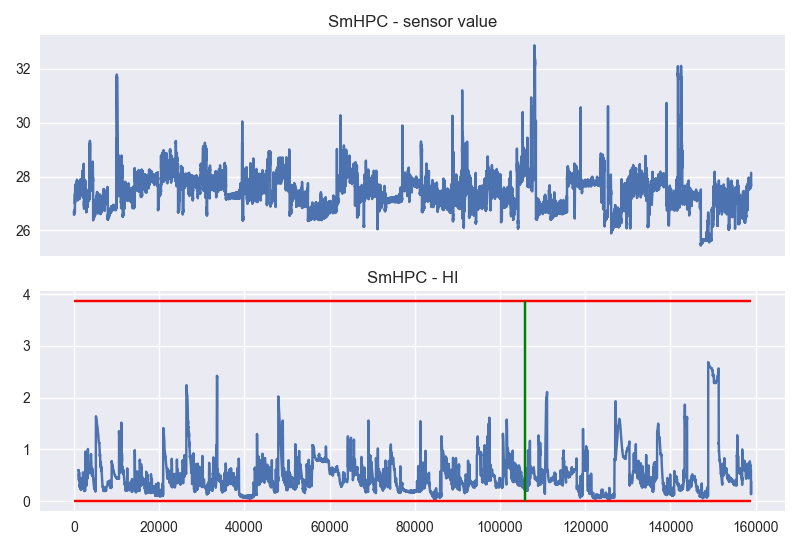}
         \caption{SmHPC sensor value and HI.}
         \label{fig:r-ncmapss-test3-tsmhpc}
     \end{subfigure}
     \hfill
     \begin{subfigure}[b]{0.45\textwidth}
         \centering
         \includegraphics[width=\textwidth]{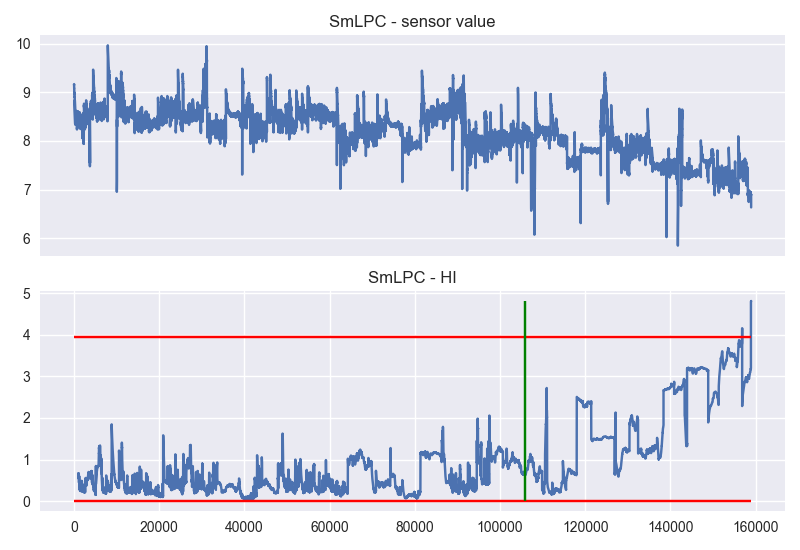}
         \caption{SmLPC sensor value and HI.}
         \label{fig:r-ncmapss-test3-smlpc}
     \end{subfigure}
     \hfill \\
     \begin{subfigure}[t]{0.45\textwidth}
         \centering
         \includegraphics[width=\textwidth]{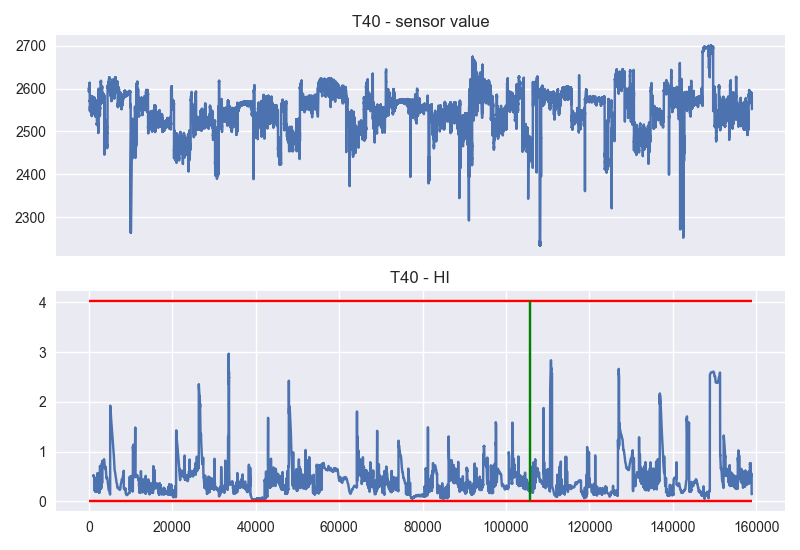}
         \caption{T40 sensor value and HI.}
         \label{fig:r-ncmapss-test3-t40}
     \end{subfigure}
     \hfill
      \begin{subfigure}[t]{0.45\textwidth}
         \centering
         \includegraphics[width=\textwidth]{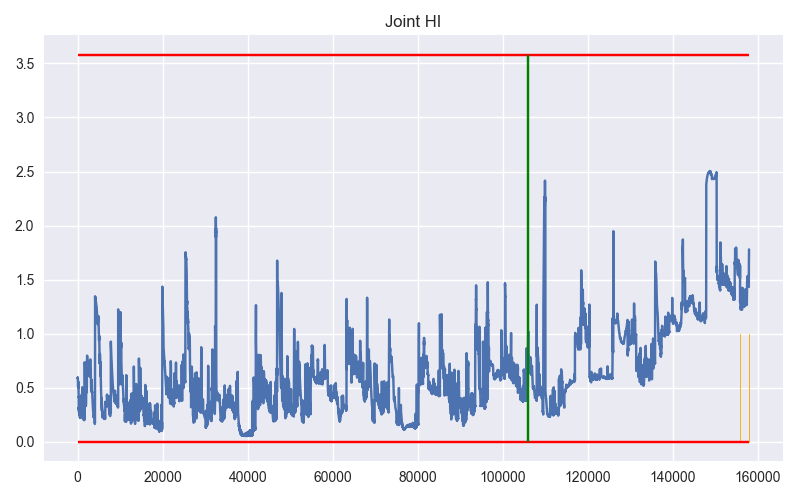}
         \caption{$\HIcombo$ for SmHPC, SMLPC, and T40.}
         \label{fig:r-ncmapss-test3-joint}
     \end{subfigure}
        \caption{The result from test 3 of the N-CMAPSS dataset. The green vertical line indicates the end of the burn-in state. The horizontal red lines indicates the calculated boundaries. The orange area in the $\HIcombo$ is when an alarm is published.}
        \label{fig:r-ncmapss-test3}
\end{figure*}

We now focus on the more challenging N-CMAPSS dataset. Here, Test 3 was designed to examine \system with multiple sensors. We chose sensors T40, SmLPC, and SmHPC, and gave them equal weight. The results in \autoref{fig:r-ncmapss-test3} show that the only component model that indicates a degradation is the SmLPC sensor. After the burn-in state, $\HImodel$ for this sensor is rising and exceeds its upper bound in the end. The same trend is also present in the \supervisor where $\HIcombo$ is approaching the upper-bound. The two alarms are published when SmLPC crosses its boundary.

\begin{figure*}
     \centering
     \begin{subfigure}[b]{0.45\textwidth}
         \centering
         \includegraphics[width=\textwidth]{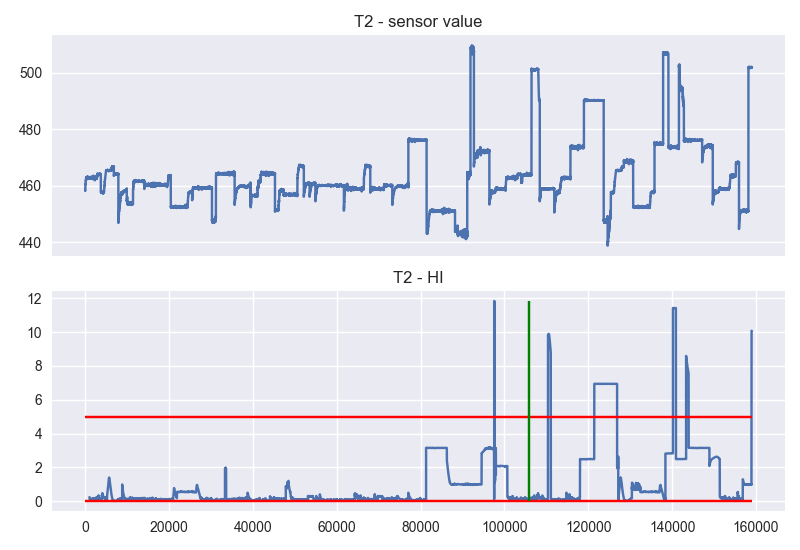}
         \caption{T2 sensor value and HI.}
         \label{fig:r-ncmapss-test5-t2}
     \end{subfigure}
     \hfill
     \begin{subfigure}[b]{0.45\textwidth}
         \centering
         \includegraphics[width=\textwidth]{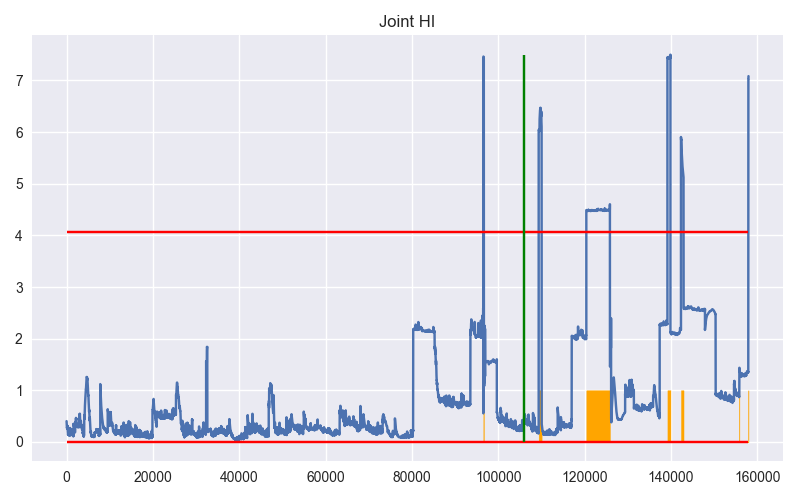}
         \caption{$\HIcombo$ for SmHPC, SmLPC, and T2.}
         \label{fig:r-ncmapss-test5-joint}
     \end{subfigure}
        \caption{The result from test 5 of the N-CMAPSS dataset. The green vertical line indicates the end of the burn-in state. The horizontal red lines indicates the calculated boundaries. The orange area in the $\HIcombo$ is when an alarm is published.}
        \label{fig:r-ncmapss-test5}
\end{figure*}

Test 5 was conducted to test \system when a higher weight is applied to one measurement, and sensor T2  was given a higher weight than the other two sensors (SmHPC and SmLPC) in this test. The T2 sensor values and its HI score can be seen in \autoref{fig:r-ncmapss-test5} along with the $\HIcombo$. Due to its higher weight, T2 highly impacts $\HIcombo$, moreover, multiple alarms are published. 

\section{Discussion} \label{sec:discussion}
The tests showed successful results on the original CMAPSS dataset. Each test was able to detect the degradation within the dataset and successfully publish an alarm at an appropriate time. \system succeeded regardless of how many non-degradation sensors were included to complicate the endeavour. This must be partially attributed to the system design, as a single \HImodel crossing its boundary is sufficient to publish an alarm. The tests on the more complex N-CMAPSS dataset were not as successful, as alarms only  were published when a single sensor value crossed its boundary. Furthermore,  sensor T2 triggered an alarm during Test 5 even though no clear degradation trend can be seen from that sensor. 

The component models used in this paper are built around the idea that one model should learn the underlying characteristics of a specific sensor. We have pre-trained our models to speed up the training process, i.e., related to the data collection and learning in the burn-in state. All the univariate models are sharing the same DAE architecture (\autoref{tab:base dae}). A direction for future research is to evaluate the performance of different architectures when applied to univariate data, potentially reflecting what sensor the data originates from. Some sensors output multivariate data streams, and while we hypothesise that a DAE architecture similar to what is currently used for univariate data could also work for multivariate streams, this is so far not thoroughly tested. Similarly, specific DAE models should also be made for images, videos, and sound to handle those types of sensor inputs. Another issue is how to model correlated sensor values. Vibration can, for instance, be measured by both a 3 degree of freedom (DoF) and 6 DoF accelerometers. To detect increasing levels of vibration it could be beneficial to analyse readings from both sensors together. The current version of \system has traded such fidelity in the model for simplicity, but the performance gains of combining correlated signals into component-overarching models should be investigated. 

The \supervisor handles all HI-values from the component models and decides when an alarm should be published. One specific challenge is that the main objective is that it should be data-driven. Therefore, the \supervisor should internally work out its optimal behaviour during the burn-in state. Currently, the \supervisor's HI value is a weighted average of those from the component models, and no computational work is done to find a better aggregation formula. Nevertheless, since we can use a \textit{weighted} average, it is possible to get at least some expert knowledge into the system by tweaking the weights. If doing so, we take a step away from an entirely data-driven approach, and move towards a hybrid approach. Another idea is to find, e.g., a neural network that uses the component models' \HImodel-values to generate \HIcombo. This would enable the \supervisor to learn the different characteristics and dependency of all the components models and thus understand the connection between sensor values, but would increase the data-requirements to initialise the system.

Currently, the boundaries from both the component models and the \supervisor are calculated by the 9'th standard deviation from the HI mean during the burn-in state. While this boundary was successful in the CMAPSS dataset as described in \autoref{sec:cmapss dataset result}, it was less successful for the N-CMAPSS dataset. This was partly because the DAE models used to calculate each component's HI value were not successful in faithfully reproducing the input-signal during the burn-in state. As a consequence, we saw a larger variability in $\HImodel$, leading to rather large standard deviation during burn-in, and therefore to a less responsive behaviour for our monitoring system. A solution could be a locally adaptive boundary that is based on the historical HI scores and changes as each new score is calculated. This would make the boundaries dynamic, but potentially also prevent the detection of slowly developing problem situations.

When working with time-series data in neural networks, the way data is presented to the model is extremely important. We use a rolling window in our implementation, and fix the window-size based on the available data and the speed of the underlying dynamics. This choice, while difficult to make, can influence the results dramatically. If the window-size is too small the model cannot learn the temporal information, and if it is too-large the result can be a prolonged training time and a reduction in available data batches reducing the learning quality. We used a window size of 8 for the CMAPSS test, and 1024 the N-CMAPSS dataset, as the dataset contained more data and high variation throughout the dataset. To reduce the sensitivity wrt. this parameter one could try to define the window size dynamically based on the amount of data collected in the burn-in state along with other statistical characteristics such as standard deviation, skewness, etc.

While other implementations of HI split them up into multiple zones compared to the severity (see, e.g., \cite{Velmurugan2021}), we have chosen to have only one type of alarm. This was done to avoid scenarios where a system would fail in, e.g. warning state instead of in a critical state. Other researchers based their zoning on historical data, which is not assumed to be present in our case. Finally, it would be beneficial for the operators to have an RUL estimation, but to the best of our knowledge there is still no method available to produce reliable RUL estimates without historical data available. 

To utilise \system in practice, a few things are considered paramount to be in place. Firstly, each component model is assumed to be already pretrained on relevant datasets. The operator will then place and select the sensor inputs through a GUI where he would define the stable period for the burn-in to occur. When the burn-in period is finished, the system will atomically go into the inference state. While we in this paper performed tests on the publicly available datasets CMAPSS and N-CMAPSS, they are not the ideal use case. \system was designed with more generic manufacturing systems in mind, such as punching machines, drill presses, CNC and bending machines. These types of systems are more comparable to the CMAPSS dataset since the system is not as complex as the data represented in the N-CMAPSS dataset. Nonetheless, as it is now, \system does not distinguish between different part numbers, e.i. \system has no way of knowing if the machine is currently producing a part it was not present in the burn-in state.
\section{Conclusion and future work} \label{sec:conclusion}
In this paper we presented \system, a novel modular architecture for constructing a HI in a production process. Our work does not require any historical data, as the DAE is fitted during a stable process on-site. Moreover, the operators choose the sensors available/required, and \system does not require any knowledge of the type of machinery. The system is data-driven, easy-to-use and modular, and these design-choices make \system suitable for SMEs that do not have the resources to invest in PHM solutions that large enterprises do.  

We tested our contribution on two open datasets, CMAPSS and N-CMAPSS. All the tests on the CMAPSS dataset successfully published an alarm and detected the degradation in the underlying process. This indicates our novel approach of using a combination multiple of individual trained DAEs aggregated together is able to produce an HI score and degradation detection without any prior historical data. This will benefit SMEs as it has been shown that they often lack the resources and knowledge to invest in large PHM solutions. 

The test of the N-CMAPSS was not as successful in detecting the degradation as on the CMAPSS tests. This is because of the higher complexity within that dataset since it contains different flights of different lengths and altitudes; state changes that influence the data substantially. We tried to overcome this problem by limiting the data to cruise flights above 10,000 feet, but found that this was not enough to make the data sufficiently stable. One could try to limit this specific dataset  to have different models for the different stages of flights. Nonetheless, this is out of scope for \system since it is focused on a more ``stable'' system, which does not change extensively throughout its execution. 

For future work, \system should be tested on real production data with various sensors to validate it in its intended settings. Moreover, tests should be carried out with both images and sound as the component models. More research should be conducted on each component model to ensure the correct architecture for each model. A more automatic system for choosing the window size should also be researched to make the system more easy-to-use. Lastly, more research within the \supervisor should be conducted to see if it can learn more from the component models instead of aggregating the results as it is now. 

\section*{Acknowledgements}
The work was conducted as a part of the Danish strategy for Industry 4.0 in SMEs called Innovation Factory North.

\bibliography{main.bib}

\end{document}

%% file: tabs/cmapss-test-table.tex
\begin{table*}[ht]
\centering
\caption{The eight test setups for CMAPSS dataset. Each sensor has its own model; if several sensors are mentioned together with only one model, then all sensors have that type of model. The same is true with the assigned weights. The model \textit{T} is pre-trained temperature data and \textit{C} is a generic model pre-trained on accelerometer data.}
\resizebox{\textwidth}{!}{%
\begin{tabular}{l|l|l|l}
\textbf{Tests} & \textbf{Sensors}                             & \textbf{Models}        & \textbf{Weights} \\ \hline
CMAPSS - T1   & T50                                          & T                   & 0.5              \\
CMAPSS - T2   & T30, T50                                     & T                  & 0.5              \\
CMAPSS - T3   & T2, T30, T50                                 & T                   & 0.5              \\
CMAPSS - T4   & T2, T30, T50                                 & T                   & 0.6, 0.2, 0.2    \\
CMAPSS - T5   & P15                                          & C                  & 0.5              \\
CMAPSS - T6   & P15, T2, T50                                 & C, T, T      & 0.5, 0.5, 0.5    \\
CMAPSS - T7   & P15, T2, T50                                 & C, T, T      & 0.2, 0.2, 0.6    \\
CMAPSS - T8   & P2, P15, epr, farB, Nf\_dmd, PCNfR\_dmd, T50 & T50: T, Rest: C & 0.5             
\end{tabular}%
}
\label{tab:cmaps test setup}
\end{table*}

%% file: tabs/ncmapss-test-table.tex
\begin{table*}[ht]
\centering
\caption{The five test setups for N-CMAPSS dataset. Each sensor has its own model; if several sensors are mentioned together with only one model, then all sensors have that type of model. The same is true with the assigned weights. The model \textit{T} is pre-trained temperature data and \textit{C} is a generic model pre-trained on accelerometer data.}
\begin{tabular}{l|l|l|l}
\textbf{Test}  & \textbf{Sensors}  & \textbf{Models}    & \textbf{Weights} \\ \hline
N-CMAPSS - T1  & T40               & T               & 0.5              \\
N-CMAPSS - T2  & SmLPC             & C              & 0.5              \\
N-CMAPSS - T3  & T40, SmLPC, SmHPC & T, C, C & 0.5              \\
N-CMAPSS - T4  & T2, SmLPC         & T, C        & 0.5              \\
N-CMAPSS - T5 & T2, SmLPC, SmHPC  & T, C, C & 0.6, 0.2, 0.2   
\end{tabular}

\label{tab:ncmaps test setup}
\end{table*}